\documentclass[letterpaper]{article}
\usepackage[preprint]{aaai2027}
\usepackage[hyphens]{url}
\usepackage{natbib}
\usepackage{graphicx}
\usepackage{caption}
\usepackage{booktabs}
\usepackage{amsmath}
\usepackage{amssymb}
\usepackage{algorithm}
\usepackage{algorithmic}
\usepackage{array}
\newcommand{\best}[1]{\textbf{\underline{#1}}}
\newcommand{\bestm}[1]{\underline{\mbox{\boldmath$#1$}}}
\title{CARGO-VL: Counterfactual Arbitration with Risk-Constrained Group Optimization for Vision--Language Models}
\author{
De Jiang\textsuperscript{\rm 1},
Zhengyang Zhang\textsuperscript{\rm 2},
Kehong Yuan\textsuperscript{\rm 3},
Shaohua Ma\textsuperscript{\rm 4}
}
\affiliations{
\small
\textsuperscript{\rm 1}\texttt{jiangd24@mails.tsinghua.edu.cn}\quad
\textsuperscript{\rm 2}\texttt{zhengyangwudi@gmail.com}\\
\textsuperscript{\rm 3}\texttt{yuankh@sz.tsinghua.edu.cn}\quad
\textsuperscript{\rm 4}\texttt{ma.shaohua@sz.tsinghua.edu.cn}
}

\begin{document}
\maketitle

\begin{abstract}
Vision--language systems combine images with retrieved text, but these sources can disagree or jointly fail to support an answer. Reliable models must identify the trustworthy source and abstain when neither is adequate. Existing post-training objectives score instances independently and therefore do not enforce coherent behavior under counterfactual evidence changes. We introduce CARGO-VL, a group-relative framework that optimizes matched variants covering aligned, image-correct, text-correct, and both-wrong (A/V/T/N) evidence states as one bundle. Its objective couples condition-wise correctness with transition rewards for answer invariance, source equivariance, and answer-to-abstention switching, while a primal--dual controller balances unsafe answers against excessive deferral. We also contribute \textbf{XMC} (e\textbf{X}tended \textbf{M}odal \textbf{C}onflict), a four-condition conflict training resource, and evaluate transfer on CMC-Bench and Modality-Bias. Across multiple seeds, CARGO-VL improves conflict handling, unsupported-answer avoidance, and modality balance over pointwise baselines. Ablations identify complementary benefits from relational transition signals and adaptive risk control, supporting counterfactual consistency as a practical objective for reliable multimodal evidence arbitration.
\end{abstract}

\section{Introduction}

Multimodal systems increasingly encounter conflicting image and textual evidence. In a product photo, chart, document, or retrieved context, one modality may be correct while the other is stale, adversarial, or irrelevant. The same tension appears in multimodal retrieval-augmented generation, where an image and a retrieved passage can both look plausible yet disagree on a date, entity, or numeric claim \citep{lewis2020rag,asai2024selfrag,catapang2026cmc}. A trustworthy model should not merely maximize answer accuracy: it should use the correct modality when one source wins and abstain when neither supports an answer. Empirically, strong vision--language models still exhibit modality preference---favoring text over image, or the reverse---even when both channels are available \citep{pezeshkpour2025mixed}, which makes conflict handling a first-class reliability problem rather than a rare corner case.

CMC-Bench makes this evaluation setting concrete through matched cross-modal evidence conflicts \citep{catapang2026cmc}.

\begin{figure}[t]
\centering
\includegraphics[width=0.98\columnwidth]{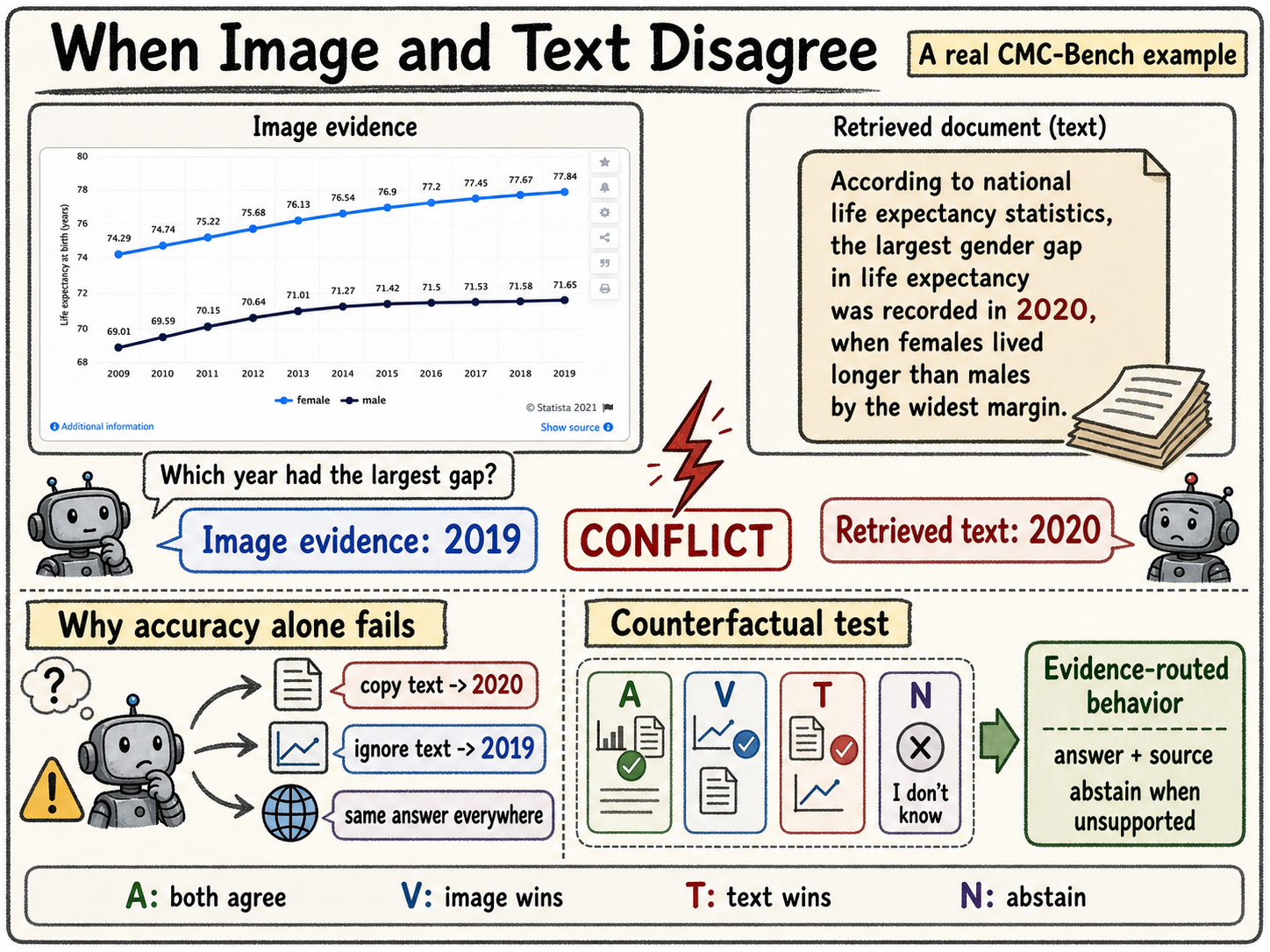}
\caption{Image--text conflict. A real CMC example supports 2019 while retrieved text claims 2020.}
\label{fig:conflict-example}
\end{figure}

\noindent It constructs aligned, image-correct, text-correct, and both-wrong (A/V/T/N) conditions and scores responses relative to the image and textual sources. Figure~\ref{fig:conflict-example} shows a temporal conflict of this form. Most multimodal benchmarks, however, still reward answering correctly under a single, fixed context. That leaves open whether the model changes its evidence attribution and final action when a controlled intervention makes a modality unreliable. The difficulty is relational: a model scored on isolated examples can obtain high average accuracy while answering the same content under all A/V/T/N conditions, citing the wrong modality when image and text disagree, or retaining an answer when both supports are invalid. Instance-wise supervised fine-tuning and standard RL post-training optimize each rollout on its own prompt \citep{ouyang2022instructgpt,schulman2017ppo,rafailov2023dpo}; they do not require coherent transitions across matched interventions of the \emph{same} question. Likewise, abstention is not monotonic---blanket refusal can look safe yet fail the task, while accuracy-only training can encourage unsupported answers---so safety and utility must be treated as competing operational costs rather than a single scalar \citep{geifman2017selective}.

These observations motivate a stricter training target. Flipping which modality is trustworthy should flip the followed source; removing both supports should flip answering to abstention; and easy aligned gains must not hide visual-wins or both-wrong failures. That requires a joint A/V/T/N optimization unit, rewards for local and relational correctness, and explicit budgets on unsafe answers and over-deferral.

We propose \textbf{C}ounterfactual \textbf{A}rbitration with \textbf{R}isk-constrained \textbf{G}roup \textbf{O}ptimization for vision--language models (\textbf{CARGO-VL}). CARGO-VL takes the matched A/V/T/N bundle as the optimization unit and scores complete bundles jointly with (i)~a strict pointwise evidence reward; (ii)~a transition reward encoding answer invariance, source equivariance, and an answer-to-abstention switch; (iii)~a soft minimum that protects the weakest condition; and (iv)~primal--dual constraints that penalize unsafe answers, excessive deferral, and degraded answer accuracy before group-relative advantage normalization. We build on GRPO \citep{shao2024deepseekmath}, changing the group from independent completions to counterfactual evidence bundles.

To avoid training on test templates, we construct \textbf{XMC} (e\textbf{X}tended \textbf{M}odal \textbf{C}onflict) by applying the A/V/T/N recipe to fresh TextVQA \citep{singh2019textvqa} and image-bearing ScienceQA \citep{lu2022scienceqa} items with new identifiers, conflict-type strata, and leakage-checked CMC-Bench holdout / Modality-Bias \citep{pezeshkpour2025mixed} evaluations. Policies emit free text; a frozen semantic judge maps ambiguous outputs to evidence labels. We evaluate with native CMC-Bench Table~3 metrics \citep{catapang2026cmc} and Mixed Signals $B=(\%\mathrm{Img}-\%\mathrm{Txt})$ \citep{pezeshkpour2025mixed}, reporting means over seeds $\{42,43,44\}$. CARGO-VL leads Table~\ref{tab:main-leaderboard}, including $B$ closest to zero; ablations reuse the same columns in Table~\ref{tab:ablation} and Figure~\ref{fig:ablation-radar}.

Our contributions are:
\begin{itemize}
    \item a bundle-level objective for evidence arbitration that scores both per-condition correctness and counterfactual transitions;
    \item a risk-constrained GRPO procedure that preserves abstention safety without collapsing answerable cases into deferral;
    \item XMC, a four-condition conflict \emph{training} set over fresh TextVQA and ScienceQA examples, with conflict-type strata and leakage-checked CMC-Bench holdout / Modality-Bias evaluations;
    \item a shared free-text evaluation protocol with a separated frozen semantic judge and multi-seed transfer evaluation on CMC-Bench holdout and Modality-Bias using their native metrics.
\end{itemize}

\section{Related Work}

\noindent\textbf{Cross-modal conflict and retrieval-conditioned generation.}
Vision--language pretraining and instruction tuning yield strong single-context answerers \citep{radford2021clip,liu2023llava}, while retrieval-augmented generation injects external evidence into the prompt \citep{lewis2020rag,asai2024selfrag}. When retrieved image and text disagree, accuracy alone is insufficient: models may lock onto a modality prior or invent unsupported answers. CMC-Bench constructs matched A/V/T/N conflicts for multimodal RAG evaluation \citep{catapang2026cmc}, and Mixed Signals measures exclusive image- versus text-favoring under vision--language conflict \citep{pezeshkpour2025mixed}. CARGO-VL targets the complementary \emph{learning} problem: optimizing a policy so that decisions change coherently across those controlled interventions, then transferring to the same native evaluation metrics.

\noindent\textbf{Post-training and counterfactual supervision.}
RLHF and preference optimization improve instruction following with instance-wise rewards or pairwise preferences \citep{ouyang2022instructgpt,rafailov2023dpo,schulman2017ppo}. GRPO replaces a learned critic with within-group relative advantages \citep{shao2024deepseekmath}. Counterfactually augmented data teaches models which features should change under interventions \citep{kaushik2020counterfactual}. CARGO-VL keeps a GRPO-style update but redefines the group as a matched evidence bundle, scoring answer invariance, source equivariance, and answer-to-abstention switching rather than treating A/V/T/N variants as independent prompts. The procedure is compatible with parameter-efficient fine-tuning, including LoRA-style adapters \citep{hu2022lora}.

\noindent\textbf{Abstention and calibrated refusal.}
Selective prediction formalizes when a model should defer rather than risk an error \citep{geifman2017selective}. In multimodal conflict, abstention must rise on both-wrong items without collapsing answerable cases into refusal. CARGO-VL therefore couples an N-switch transition with primal--dual penalties on unsafe answers, over-deferral, and supported exact-match loss, rather than a single accuracy--refusal scalar.

\section{CARGO-VL Training Strategy}

\begin{figure*}[t]
\centering
\includegraphics[width=0.98\textwidth]{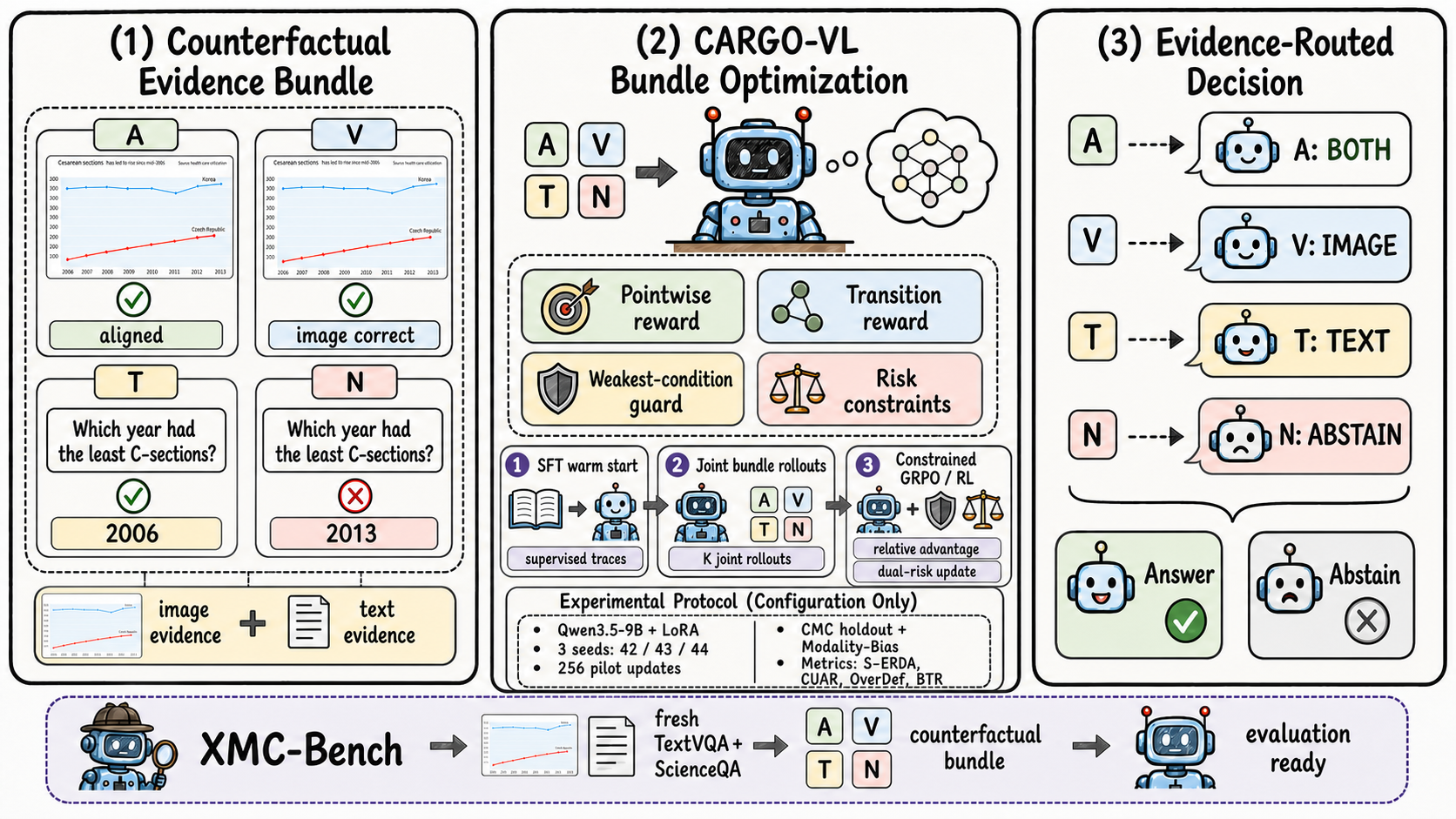}
\caption{CARGO-VL overview. A real CMC-Bench ChartQA input illustrates how the same question is instantiated as an aligned (A), image-correct (V), text-correct (T), or both-wrong (N) evidence bundle. CARGO-VL jointly scores all four conditions with pointwise and transition rewards, weakest-condition protection, and risk constraints, then optimizes the policy with an SFT warm start followed by constrained GRPO. The protocol panel records the training configuration; it reports no performance values. The XMC training set extends the four-condition recipe to fresh TextVQA and ScienceQA examples used only for training.}
\label{fig:cargo-overview}
\end{figure*}

\subsection{Problem Setup}

For a base question $q$ and image $x$, let $B=\{s_A,s_V,s_T,s_N\}$ be a matched bundle. Conditions A, V, T, and N respectively denote aligned evidence, image-correct evidence, text-correct evidence, and both-wrong evidence. Each sample provides a gold answer $y$ for answerable conditions and a target action: answer for A/V/T and abstain for N. A policy $\pi_\theta$ emits free text $o_c$ for each condition $c\in B$. A frozen semantic mapper maps $o_c$ to a decision $d_c\in\{\textsc{answer},\textsc{deflect}\}$ and source $z_c\in\{\textsc{image},\textsc{text},\textsc{both},\textsc{none}\}$.

The desired bundle behavior is simple but demanding: A, V, and T should answer $y$; their sources should be both, image, and text; and N should abstain with source none. The matched construction removes a confound present in a random grouping of unrelated questions: every transition is defined with respect to the same underlying question and answer. Table~\ref{tab:behavior} makes the behavioral contract explicit. An answer in V is insufficient if the model attributes the answer to text, and a correct-looking string in N is insufficient if the model declines to abstain. During training, each condition uses the same free-text interface; source and decision fields come from the semantic mapping layer rather than brittle structured policy outputs.

\begin{table}[t]
\centering
\small
\begin{tabular}{@{}c p{0.31\columnwidth} p{0.22\columnwidth} p{0.20\columnwidth}@{}}
\toprule
Cond. & Evidence state & Required action & Required source \\
\midrule
A & image and text agree & answer $y$ & both \\
V & image correct, text conflicts & answer $y$ & image \\
T & text correct, image conflicts & answer $y$ & text \\
N & neither source supports $y$ & abstain & none \\
\bottomrule
\end{tabular}
\caption{Target behavior of a complete counterfactual evidence bundle.}
\label{tab:behavior}
\end{table}

\subsection{Condition-Level Reward}

For condition $c$, free text $o_c$ is mapped to $\ell_c\in\{\textsc{image},\textsc{text},\textsc{both},\textsc{neither},\textsc{abstain}\}$ by a frozen mapper (exact/numeric/refusal rules, else a frozen LLM judge) that never sees gold $y$. From $\ell_c$ we set decision $d_c$ and source $z_c$ (\textsc{abstain}${\to}$deflect/none; \textsc{neither}${\to}$answer/none). With targets $\ell_c^\star$ for A/V/T/N and weights $(w_{\mathrm{act}},w_{\mathrm{src}},w_{\mathrm{ans}})=(1.0,0.8,1.0)$,
\begin{align}
r_{\mathrm{act}}&=\mathrm{I}[d_c{=}d_c^\star],\;
r_{\mathrm{src}}=\mathrm{I}[\ell_c{=}\ell_c^\star],\nonumber\\
r_{\mathrm{ans}}&=\mathrm{I}[\mathrm{ans.}(c)\land\mathrm{match}(o_c,y)],
\end{align}
with $\mathrm{ans.}(c)$ denoting $c{\in}\{\mathrm{A},\mathrm{V},\mathrm{T}\}$ and $d_c{=}\textsc{answer}$,
where $d^\star$ is answer on A/V/T and deflect on N. So $r_{\mathrm{act}}$/$r_{\mathrm{src}}$ use mapper outputs only; $r_{\mathrm{ans}}$ uses gold only via post-mapping string match. Then
\begin{equation}
r_{\mathrm{abs}}=
\begin{cases}
w_{\mathrm{act}}r_{\mathrm{act}}, & \mathrm{N},\\
w_{\mathrm{act}}r_{\mathrm{act}}+w_{\mathrm{src}}r_{\mathrm{src}}+w_{\mathrm{ans}}r_{\mathrm{ans}}, & \mathrm{A/V/T}.
\end{cases}
\end{equation}
On N only \textsc{abstain} scores; on A/V/T, wrong source loses $r_{\mathrm{src}}$ and wrong string loses $r_{\mathrm{ans}}$.

\subsection{Bundle Transition Reward}

Pointwise correctness leaves the relationship among variants unconstrained. Let $p_A,p_V,p_T,p_N$ be mapped free-text answers from one joint rollout, $\mathbf d=(d_A,d_V,d_T,d_N)$ its decision vector, and $(s_A^\star,s_V^\star,s_T^\star)=(\textsc{both},\textsc{image},\textsc{text})$ the supported-condition source targets. Let $\mathbf d^\star=(\textsc{answer},\textsc{answer},\textsc{answer},\textsc{deflect})$. We define three transition components:
\begin{align}
r_{\mathrm{AI}} &= \mathrm{I}\!\Big[{\textstyle\bigwedge_{c\in\{A,V,T\}}}
\bigl(d_c{=}\textsc{answer}\land\mathrm{match}(p_c,y)\bigr)\Big],\nonumber\\
r_{\mathrm{SE}} &= \tfrac{1}{3}\textstyle\sum_{c\in\{A,V,T\}}\mathrm{I}[z_c=s_c^\star],\nonumber\\
r_{\mathrm{DS}} &= \mathrm{I}\big[\mathbf d=\mathbf d^\star\ \land\ z_N=\textsc{none}\big],
\end{align}
and $r_{\mathrm{trans}}=0.4r_{\mathrm{AI}}+0.3r_{\mathrm{SE}}+0.3r_{\mathrm{DS}}$. Thus, correctness under a modality perturbation is not enough: the model must identify the winning modality, and it must switch from answering to deferring once both supports are invalid.

For one bundle rollout, the total unpenalized reward is
\begin{equation}
\begin{aligned}
R_B={}&\tfrac{1}{4}\textstyle\sum_{c\in B}r_{\mathrm{abs}}(c)\\
&+\eta r_{\mathrm{trans}}+\beta\,\operatorname{softmin}_{\tau}\!\big(\{r_{\mathrm{abs}}(c)\}_{c\in B}\big),
\end{aligned}
\end{equation}
with $(\eta,\beta,\tau)=(0.5,0.3,0.2)$ and $\operatorname{softmin}_{\tau}(u)=-\tau\log\bigl(\frac{1}{|u|}\sum_i\exp(-u_i/\tau)\bigr)$. The soft minimum reduces the incentive to trade a weak counterfactual condition for a strong average.

\subsection{Combining Local and Relational Signals}

The pointwise terms make every response locally accountable---action, source, and answer string---but do not require a coherent shift from source \textsc{both} to \textsc{image} or \textsc{text} when the supported modality flips. The transition term supplies that relational signal; the soft-min protects the weakest condition in the bundle. Removing a component therefore predicts a specific regression (local mismatch, inconsistent counterfactual behavior, or uneven robustness) rather than an undifferentiated score change.

\subsection{Adaptive Risk-Constrained Optimization}

Safety rewards can be neutralized by group normalization if they are batch constants. CARGO-VL therefore applies rollout-level costs before computing relative advantages. For dual variables $\lambda=(\lambda_u,\lambda_o,\lambda_e)$, each condition contributes an indicator-style cost centered at its budget: unsafe answers on deflectable items ($c{=}\mathrm{N}$), deferrals on answerable items, and supported exact-match misses. Averaging over the four conditions yields $C_B(\lambda)$; the constrained score is $\widetilde R_B=R_B-C_B(\lambda)$. For $K$ jointly sampled rollouts of a bundle, we normalize $\widetilde R_B$ within the group and perform a GRPO-style update \citep{shao2024deepseekmath}.

After every fourth policy update, projected dual ascent uses an independent stratified risk batch:
\begin{equation}
\begin{aligned}
\lambda_u &\leftarrow \Pi_{[0,\lambda_{\max}]}\!\left[\lambda_u+\rho(\widehat C_u-\varepsilon_u)\right],\\
\lambda_o &\leftarrow \Pi_{[0,\lambda_{\max}]}\!\left[\lambda_o+\rho(\widehat C_o-\varepsilon_o)\right],\\
\lambda_e &\leftarrow \Pi_{[0,\lambda_{\max}]}\!\left[\lambda_e+\rho(\varepsilon_e-\widehat{\mathrm{EM}})\right],
\end{aligned}
\end{equation}
with $(\varepsilon_u,\varepsilon_o)=(0.35,0.10)$, $\varepsilon_e=\mathrm{EM}_{\mathrm{base}}-0.05$ from the SFT baseline, $\rho=0.05$, and $\lambda_{\max}=10$. The risk batch is not reused as the policy-gradient batch. This design makes abstention a measurable constraint rather than an unconstrained reward heuristic.

\subsection{Implementation Details and Guardrails}

The implementation groups examples by a common base identifier and rejects incomplete bundles for the relational reward. It explicitly disallows a historical random-bundle control because unrelated questions do not share a valid transition gold target. For each selected bundle, the trainer samples $K{=}2$ joint completions, computes per-condition pointwise values, evaluates the transition relation, and computes one constrained reward per joint rollout. The advantage is normalized only after the individual cost has been applied.

Resolved reward settings, dual-state snapshots, and run metadata are logged per seed. The shared configuration uses Qwen3.5-9B with LoRA rank $32$ / $\alpha{=}64$, learning rate $5{\times}10^{-6}$, $K{=}2$ joint rollouts per update, $256$ RL updates, KL coefficient $0.02$, and a size-$16$ stratified risk batch, with seeds $\{42,43,44\}$. The static-penalty ablation disables dual ascent and instead subtracts fixed costs $1.5$ (unsafe answer on N) and $0.5$ (over-deferral on A/V/T), averaged over the four conditions. These settings are fixed protocol values, not tuned claims of optimality.

\begin{algorithm}[t]
\caption{CARGO-VL bundle update}
\label{alg:cargo}
\begin{algorithmic}[1]
\REQUIRE Complete A/V/T/N bundles, policy $\pi_\theta$, dual state $\lambda$
\FOR{each update}
  \STATE Sample matched bundles and $K$ joint free-text rollouts per bundle.
  \STATE Map outputs to semantic decisions and sources using the frozen labeler.
  \STATE Compute $R_B$ from pointwise, transition, and soft-min terms.
  \STATE Subtract rollout-level risk cost $C_B(\lambda)$ before group normalization.
  \STATE Update $\theta$ with group-relative advantages.
  \STATE Score an independent stratified risk batch and project the dual update.
\ENDFOR
\end{algorithmic}
\end{algorithm}

\section{Experiments}

\subsection{Benchmark Construction and Evaluation Interface}

The training and evaluation pipeline builds a common schema across CMC evaluation bundles and the XMC training set. An example is eligible for the transition objective only when all A/V/T/N variants are present, their base identifiers agree, and modality-specific references satisfy the expected relation. In particular, A and N require agreeing image and text references; V and T require a conflict; and the answerable modality must match the gold answer in A, V, or T. These checks are performed before reward computation so that missing or contradictory metadata cannot silently create a spurious relational target.

\subsection{Free-Text Semantic Mapping}

The policy is trained and evaluated as a free-text generator. It is not asked to emit JSON, action tokens, or a source field. Instead, a frozen mapper returns a five-way label. Deterministic rules are used only for high-confidence exact, numeric, and explicit-refusal cases; ambiguous outputs are passed to the frozen judge. The tested policy is unloaded before the judge is loaded, which makes the inference boundary and device budget explicit. We release the judge prompt, model revision, decoding settings, cache key, and the count of deterministic versus judge-resolved labels.

As above, the mapper never receives an answer key when assigning $\ell_c$; gold $y$ enters only $r_{\mathrm{ans}}$ and the transition answer-invariance term via $\mathrm{match}(\cdot,y)$. This separation prevents the evaluator from conflating evidence routing with answer correctness.

\subsection{Experimental Details}

\subsubsection{Data and Models}

Policies are trained under the XMC dual-bench protocol and evaluated on CMC-Bench holdout and Modality-Bias. Native XMC applies the CMC A/V/T/N recipe to fresh TextVQA and image-bearing ScienceQA validation items \citep{singh2019textvqa,lu2022scienceqa} with same-type distractors, distinct same-pool images, and new identifiers; only complete bundles with agreeing A/N references, conflicting V/T evidence, and gold-consistent answerable modalities are kept (invalid references dropped). Train mixes $627$ matched bundles ($2{,}508$ rows; $233$ native~$+$~$394$ CMC non-holdout) over temporal/factual/entity ($200$/$214$/$213$), holds out granularity as template OOD ($234$ bundles), and never uses Modality-Bias. Dev/test hold $74$/$123$ bundles; primary tests are CMC-Bench holdout ($283$/$1{,}132$; ${\approx}30\%$ template-stratified) and Modality-Bias VSR ($846$). Leakage checks report disjoint CMC holdout \texttt{base\_id}s from train/dev/test/ood, no Modality-Bias image/question overlap with training, template-disjoint train/OOD, and versus CMC holdout $0$ base-id / $2$ question-string / $91$ image-pool-path overlaps (pool reuse without instance-id leakage). The policy is Qwen3.5-9B with LoRA rank~$32$ / $\alpha{=}64$, learning rate $5{\times}10^{-6}$, $K{=}2$, $256$ RL updates, and seeds $\{42,43,44\}$.

All policy outputs are free text. A frozen Qwen3.5-27B judge maps only ambiguous outputs to \{\textsc{image}, \textsc{text}, \textsc{both}, \textsc{neither}, \textsc{abstain}\}; high-confidence exact, numeric, and refusal cases may be resolved deterministically. The policy is unloaded before the judge is loaded.

\subsubsection{Metrics and Comparisons}

The CMC-Bench holdout experiment uses the full official CMC-Bench Table~3 suite \citep{catapang2026cmc}: per-condition Acc($c$) from judge labels (Acc(A)/Acc(I$^\star$)/Acc(T$^\star$)/Acc(N)), modality preference bias (MPB-img/txt among IMAGE/TEXT commits), modality-following rate (MFR) on conflict conditions, ConfabR and CDR (NEITHER and ABSTAIN rates on conflict trials), and $\Delta\mathrm{Acc}=\mathrm{Acc}(\mathrm{aligned})-\mathrm{mean}\,\mathrm{Acc}(\mathrm{conflict})$. We also list diagnostic $\mathrm{HR}=\mathrm{ConfabR}+\mathrm{CDR}$ but do not rank by HR, because ConfabR is minimized while CDR is maximized. Official Acc(N) counts both \textsc{neither} and \textsc{abstain} as hits, while training $r_{\mathrm{act}}$ on N credits only \textsc{abstain}. Held-out Modality-Bias uses the Mixed Signals bias $B=(\%\mathrm{Img}-\%\mathrm{Txt})$ \citep{pezeshkpour2025mixed}. For GPT-4o we report only the published VSR $B$ from Mixed Signals Table~4; Table~3 single-modality accuracies and mitigation Acc/F1 are a different protocol and are not substituted for $\%$Img/$\%$Txt.

Table~\ref{tab:fair-protocol} states the fair comparison protocol. SFT/GRPO/CFPO/CARGO-VL share XMC train data, SFT init, LoRA, decoding, $256$ RL updates ($3072$ matched completions), and the free-text judge. \textbf{GRPO} optimizes pointwise $r_{\mathrm{abs}}$ only \citep{shao2024deepseekmath}. \textbf{CFPO} adds a visual factual/counterfactual sensitivity bonus (weight $0.5$) \citep{kaushik2020counterfactual} without CARGO's full A/V/T/N transition or duals. Base/Gemma-26B/GPT-4o are eval-only. Table~\ref{tab:main-leaderboard} reports means over seeds $\{42,43,44\}$.

\begin{table}[t]
\centering
\small
\setlength{\tabcolsep}{3pt}
\begin{tabular}{@{}lcccc@{}}
\toprule
 & SFT & GRPO & CFPO & CARGO-VL \\
\midrule
Data/init & XMC/-- & XMC/SFT & XMC/SFT & XMC/SFT \\
LoRA & 32/64 & 32/64 & 32/64 & 32/64 \\
LR/KL & -- & $5\mathrm{e}{-}6$/0.02 & same & same \\
RL budget & -- & 256 (3072) & 256 (3072) & 256 (3072) \\
Objective & CE & pointwise & +vis.\ CF & +trans.+dual \\
Seeds & $\{42,43,44\}$ & same & same & same \\
\bottomrule
\end{tabular}%
\caption{Fair comparison protocol (shared free-text judge). CFPO adds visual CF sensitivity; only CARGO-VL uses transition rewards and projected duals.}
\label{tab:fair-protocol}
\end{table}

\subsection{Main Results on CMC-Bench holdout and Modality-Bias}

\subsubsection{CMC-Bench holdout and Modality-Bias Leaderboard}

Table~\ref{tab:main-leaderboard} reports both native evaluation suites. The CMC-Bench holdout experiment evaluates A/V/T/N routing under retrieved image--text conflict, while Modality-Bias reports Mixed Signals bias $B$ on the VSR split.

\begin{table*}[t]
\centering
\small
\setlength{\tabcolsep}{2.6pt}
\begin{tabular}{@{}lccccccccccc|ccc@{}}
\toprule
& \multicolumn{11}{c|}{\textbf{CMC-Bench holdout}} & \multicolumn{3}{c}{\textbf{Modality-Bias}} \\
\cmidrule(lr){2-12}\cmidrule(l){13-15}
Method & Acc(A)$\uparrow$ & Acc(I$^\star$)$\uparrow$ & Acc(T$^\star$)$\uparrow$ & Acc(N)$\uparrow$ & MPB$_\mathrm{i}$ & MPB$_\mathrm{t}$ & MFR$\uparrow$ & Conf.$\downarrow$ & CDR$\uparrow$ & HR$^\ddagger$ & $\Delta$Acc$\downarrow$ & $B$($\to$0) & $\%$Img & $\%$Txt \\
\midrule
Base & 93.8 & 41.8 & 89.5 & 85.0 & 44.0 & 56.0 & 58.5 & 12.0 & 26.5 & 38.5 & 21.7 & $-7.8$ & 3.8 & 11.6 \\
\midrule
SFT & 98.7 & 43.8 & 99.6 & 92.2 & 42.0 & 58.0 & 79.4 & 18.5 & 28.4 & 46.9 & 20.2 & $+7.8$ & 12.6 & 4.8 \\
GRPO & 99.0 & 43.1 & 99.6 & 96.8 & 45.0 & 55.0 & \best{80.0} & 13.7 & 36.4 & 50.1 & 19.2 & $+6.1$ & 9.1 & 3.0 \\
CFPO & \best{99.3} & 43.5 & 99.3 & 95.1 & 47.0 & 53.0 & 79.5 & 12.8 & 38.3 & 51.1 & 20.0 & $+5.8$ & 11.5 & 5.7 \\
\midrule
Gemma-26B & 95.2 & 43.0 & 91.0 & 87.5 & 46.0 & 54.0 & 63.5 & 10.2 & 31.0 & 41.2 & 21.4 & $+4.5$ & 16.5 & 12.0 \\
GPT-4o$^\dagger$ & 94.7 & 44.5 & 91.2 & 88.3 & 33.0 & 67.0 & 71.6 & 14.8 & 32.0 & 46.8 & 20.0 & $+6.9$ & --- & --- \\
\midrule
CARGO-VL & 99.1 & \best{48.1} & \best{100.0} & \best{98.2} & \best{49.0} & \best{51.0} & 79.6 & \best{7.3} & \best{53.2} & 60.5 & \best{17.0} & \bestm{-3.2} & \best{29.2} & \best{32.4} \\
\bottomrule
\end{tabular}
\caption{Main leaderboard: CMC-Bench holdout / Modality-Bias (mean over seeds $\{42,43,44\}$). Left block matches the official CMC-Bench Table~3 suite \citep{catapang2026cmc}: Acc($c$) from judge labels; $\mathrm{MPB}_\mathrm{img}=N_\mathrm{img}/(N_\mathrm{img}+N_\mathrm{txt})$ and $\mathrm{MPB}_\mathrm{txt}=1-\mathrm{MPB}_\mathrm{img}$ among single-modality commits (values ${>}50$ indicate a systematic lean); MFR on conflict conditions; ConfabR/CDR $=$ NEITHER/ABSTAIN on conflict trials; diagnostic $\mathrm{HR}=\mathrm{ConfabR}+\mathrm{CDR}$ (no preferred direction: ConfabR$\downarrow$ while CDR$\uparrow$); $\Delta\mathrm{Acc}=\mathrm{Acc}(\mathrm{aligned})-\mathrm{mean}\,\mathrm{Acc}(\mathrm{conflict})$. Right block matches Mixed Signals bias reporting \citep{pezeshkpour2025mixed}: $B=(\%\mathrm{Img}-\%\mathrm{Txt})$. Higher is better except ConfabR and $\Delta$Acc; for $B$ and MPB, closer to zero / $50$--$50$ is better; HR is not ranked. Best values are marked in bold with underline. Ablations reuse these columns in Table~\ref{tab:ablation} and Figure~\ref{fig:ablation-radar}.}
\label{tab:main-leaderboard}

{\footnotesize $^\ddagger$HR${=}$ConfabR${+}$CDR is reported only as a CMC diagnostic and is not used for ranking (ConfabR and CDR should be read separately). $^\dagger$Modality-Bias for GPT-4o reports only the published Mixed Signals \emph{VSR} bias $B{=}{+}6.9$ (Table~4; $n{=}846$) \citep{pezeshkpour2025mixed}. Mixed Signals also lists GPT-4o VSR single-modality / aligned accuracies (Table~3: Only-Image $67.7$, Only-Text $98.8$, Pair $86.1$) and mitigation mismatch scores; those are \emph{not} our $\%$Img/$\%$Txt columns (exclusive favoring rates under our free-text protocol) and are not copied here. GPT-4o $B$ is task-dependent in Mixed Signals (e.g., Connectivity $+52.3$, Convexity $-65.2$); we use the VSR entry because it matches our Modality-Bias evaluation. CMC columns for GPT-4o are under our judge protocol (CMC-Bench does not evaluate GPT-4o). Conf.\ $=$ ConfabR.}
\end{table*}

\begin{figure*}[t]
\centering
\includegraphics[width=\textwidth,height=1.35in,keepaspectratio]{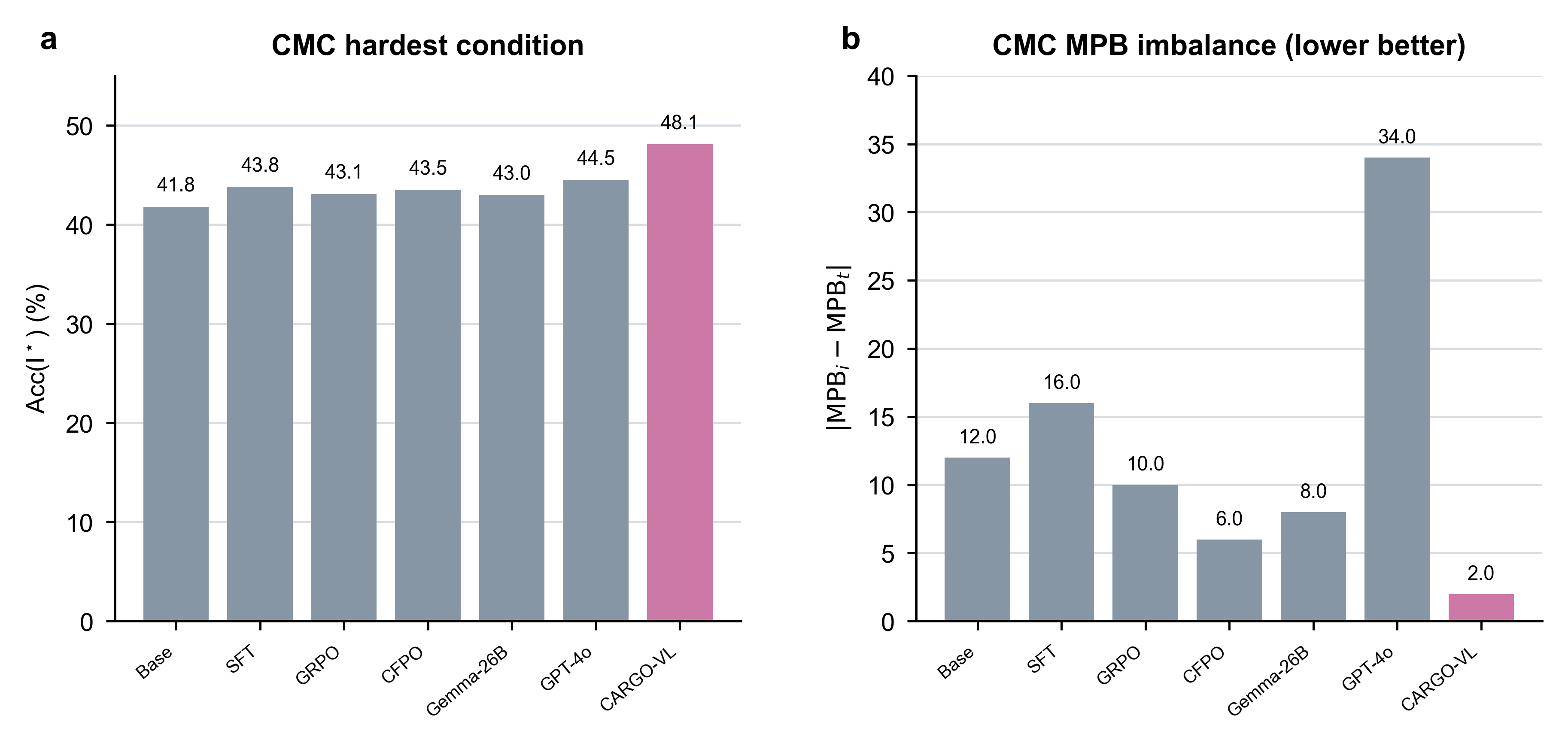}
\caption{CMC-Bench holdout / Modality-Bias diagnostics beyond the raw table cells. a, CMC Acc(I$^\star$) (hardest condition; 1,132 examples). b, CMC modality-preference imbalance $|\mathrm{MPB}_\mathrm{i}-\mathrm{MPB}_\mathrm{t}|$ (lower is closer to a $50$--$50$ split). Magenta denotes CARGO-VL. Means over seeds $\{42,43,44\}$ under the shared free-text judge protocol.}
\label{fig:benchmark-results}
\end{figure*}

On CMC-Bench holdout, Acc(A) and Acc(T$^\star$) are already strong for fine-tuned methods, so ranking is driven by Acc(I$^\star$), Acc(N), ConfabR, CDR, and $\Delta$Acc. CARGO-VL leads these axes (Acc(I$^\star$)${=}48.1$, Acc(N)${=}98.2$, ConfabR${=}7.3$, CDR${=}53.2$, $\Delta$Acc${=}17.0$), while Acc(A)${=}99.1$ and MFR${=}79.6$ trail CFPO ($99.3$) and GRPO ($80.0$) slightly. The Acc(I$^\star$) margin over GPT-4o ($44.5$) and SFT ($43.8$) is the clearest evidence that bundle-level source equivariance helps on visual-wins conflict, where instance-wise objectives tend to lock onto text. External references sit between Base and the fine-tuned group on most CMC columns (Gemma-26B Acc(I$^\star$)${=}43.0$), confirming that scale alone does not solve the problem. Diagnostic HR is listed for CMC compatibility but is not ranked, because ConfabR and CDR move in opposite preferred directions.

On Modality-Bias, CARGO-VL attains $B{=}{-}3.2$, the closest to zero among methods with measured exclusive $\%$Img/$\%$Txt, while raising both exclusive rates ($\%$Img${=}29.2$, $\%$Txt${=}32.4$) rather than collapsing into blanket abstention. GPT-4o contributes only the published VSR $B{=}{+}6.9$ from Mixed Signals Table~4; its sign of $B$ flips across Mixed Signals tasks, so $+6.9$ is a VSR-specific transfer point rather than a universal image prior. Figure~\ref{fig:benchmark-results} highlights Acc(I$^\star$) and the CMC MPB gap $|\mathrm{MPB}_\mathrm{i}-\mathrm{MPB}_\mathrm{t}|$, where CARGO-VL is the only method near a $50$--$50$ commit split.

\begin{figure*}[t]
\centering
\small
\setlength{\tabcolsep}{3.2pt}
\begin{tabular}{@{}lccccccc@{}}
\toprule
Method & Acc(I$^\star$) & Acc(N) & Conf. & CDR & $\Delta$Acc & $B$ & $\%$Img \\
\midrule
w/o transition & 44.9 & 95.4 & 22.4 & 26.4 & 19.6 & $+12.9$ & 27.4 \\
static penalty & 44.9 & 94.0 & 19.6 & 28.5 & 20.2 & $+13.5$ & 24.0 \\
CARGO-VL & \best{48.1} & \best{98.2} & \best{7.3} & \best{53.2} & \best{17.0} & \bestm{-3.2} & \best{29.2} \\
\bottomrule
\end{tabular}
\captionof{table}{Ablations on CMC-Bench holdout / Modality-Bias (mean over seeds $\{42,43,44\}$). CARGO-VL numbers match Table~\ref{tab:main-leaderboard}.}
\label{tab:ablation}

\includegraphics[width=0.30\textwidth]{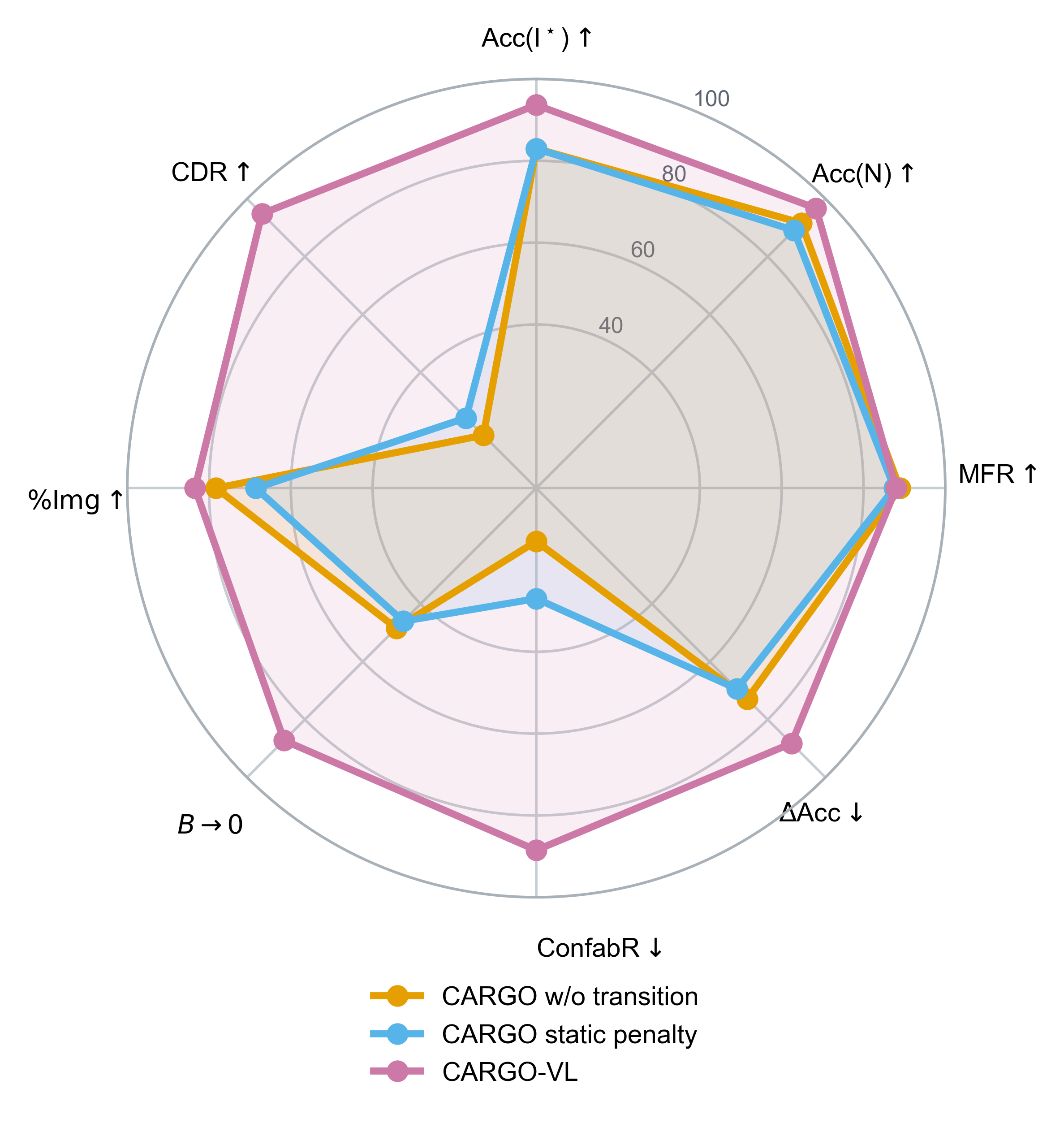}
\captionof{figure}{Transition reward and adaptive risk control. Radar of Table~\ref{tab:ablation}; outward better after reversing $\Delta$Acc, ConfabR, and $|B|$.}
\label{fig:ablation-radar}
\end{figure*}

\subsubsection{Differences in Modality Evidence Trust}

We next use Table~\ref{tab:main-leaderboard} to compare how methods \emph{credit} image versus text when the two disagree. Three complementary views are available: (i)~condition accuracy Acc(I$^\star$) versus Acc(T$^\star$), which asks whether the model follows the gold-supported modality; (ii)~CMC MPB, which records the IMAGE/TEXT share among single-modality commits; and (iii)~Modality-Bias $B$ with exclusive $\%$Img/$\%$Txt on the VSR conflict set.

\noindent\textbf{Static text lean versus calibrated routing.}
Base favors text on CMC (MPB $44.0$/$56.0$), with Acc(T$^\star$)${=}89.5$ far above Acc(I$^\star$)${=}41.8$. SFT raises answer accuracy but retains an asymmetric routing profile (MPB $42.0$/$58.0$; $B{=}{+}7.8$). GRPO and CFPO reduce the CMC MPB gap to $10$ and $6$, respectively, yet Acc(I$^\star$) remains near $43$--$44$. GPT-4o has the strongest CMC text lean (MPB $33.0$/$67.0$), whereas its published VSR $B{=}{+}6.9$ is comparatively mild, underscoring that bias is task-dependent.

\noindent\textbf{CARGO-VL reduces preference without sacrificing the weaker modality.}
CARGO-VL jointly reduces the CMC MPB gap to $2$, raises Acc(I$^\star$) to $48.1$ while retaining Acc(T$^\star$)${=}100.0$, and reaches $B{=}{-}3.2$ with substantial exclusive image/text rates ($29.2/32.4$). Thus, near-zero bias is not obtained through blanket refusal. Relative to SFT, ConfabR falls from $18.5$ to $7.3$ and CDR rises from $28.4$ to $53.2$, while MFR${=}79.6$ remains close to GRPO ($80.0$). The MPB-gap sequence $16{\rightarrow}10{\rightarrow}6{\rightarrow}2$ from SFT through CARGO-VL supports the role of source-equivariant transitions and the N-switch beyond pointwise rewards alone.

\subsection{Ablation Study}

We ablate two CARGO-specific components while holding the XMC training split, LoRA policy, update budget, seeds $\{42,43,44\}$, and free-text judge fixed: (i)~removing the bundle transition reward (pointwise $+$ soft-min only), and (ii)~replacing projected dual penalties with fixed static costs. Table~\ref{tab:ablation} and Figure~\ref{fig:ablation-radar} use the same native columns as Table~\ref{tab:main-leaderboard}.

\noindent\textbf{Transition reward drives source routing.}
Dropping $r_{\mathrm{trans}}$ cuts Acc(I$^\star$) from $48.1$ to $44.9$ and Acc(N) from $98.2$ to $95.4$, while MFR stays nearly flat ($79.6{\rightarrow}80.0$). The sharper failures are relational: ConfabR rises $7.3{\rightarrow}22.4$ and CDR falls $53.2{\rightarrow}26.4$, so the policy answers unsupported conflict items more often and detects conflict less often. On Modality-Bias, $B$ moves from $-3.2$ to $+12.9$ with $\%$Img$/\%$Txt $=27.4/14.5$, reintroducing a one-sided favor rate. CMC MPB likewise regresses from $49.0$/$51.0$ to $30.1$/$69.9$. These shifts match the transition terms (source equivariance and N-switch): without them, pointwise $r_{\mathrm{act}}$/$r_{\mathrm{src}}$/$r_{\mathrm{ans}}$ recover GRPO-like conflict behavior rather than bundle-consistent routing.

\noindent\textbf{Static penalties under-calibrate abstention.}
Replacing dual updates with fixed costs yields the same Acc(I$^\star$)${=}44.9$ but weaker safety--utility balance: Acc(N)${=}94.0$, ConfabR${=}19.6$, CDR${=}28.5$, and $\Delta$Acc${=}20.2$. Modality-Bias $B{=}{+}13.5$ with $\%$Img${=}24.0$ and $\%$Txt${=}10.5$, and the Modality-Bias abstain rate rises to $42.0$ (versus $1.1$ for full CARGO-VL), indicating over-deferral under a non-adaptive multiplier. Projected duals are therefore not redundant with the transition reward: they keep refusal high where evidence is unsupported while preserving exclusive modality rates closer to parity.

\noindent\textbf{Complementary contributions.}
Full CARGO-VL is outermost on the radar after flipping lower-is-better axes, jointly improving Acc(I$^\star$), Acc(N), ConfabR, CDR, $\Delta$Acc, and $|B|$. Transition consistency supplies routing relative to SFT/GRPO; projected duals prevent confabulation and blanket abstention.

\noindent\textbf{Joint interpretation and scope.}
Both reduced variants reach Acc(I$^\star$)${=}44.9$, $3.2$ points below full CARGO-VL, but their remaining error profiles differ. Relative to the transition-free variant, the full objective lowers ConfabR by $15.1$ points and raises CDR by $26.8$ points; relative to static penalties, it lowers ConfabR by $12.3$ points and raises CDR by $24.7$ points. Both ablations also produce $B{>}{+}12$, whereas full CARGO-VL reaches $-3.2$ while retaining substantial exclusive image/text rates. These joint movements support complementary roles for relational routing and adaptive calibration under the fixed XMC split, backbone, update budget, and judge protocol. They do not establish that the selected transition weights, soft-min temperature, or risk budgets are individually optimal.

\noindent\textbf{Matched visual accuracy does not imply matched conflict behavior.}
The identical Acc(I$^\star$)${=}44.9$ of both reduced variants controls for raw visual-answer capacity. Their divergent Acc(N), ConfabR, and $\Delta$Acc therefore expose different conflict failures, while full CARGO-VL improves all four quantities rather than exchanging visual routing for abstention.

\noindent\textbf{Bias must be interpreted with decision coverage.}
A near-zero bias can be misleading when a policy avoids modality-specific decisions. Full CARGO-VL combines $B{=}{-}3.2$ with exclusive image/text rates of $29.2/32.4$, whereas static penalties yield $B{=}{+}13.5$, rates of $24.0/10.5$, and a $42.0$ abstain rate. Joint reporting therefore distinguishes balanced routing from low-coverage behavior and evaluates the dual controller on both safety and utility.

\section{Conclusion}

CARGO-VL optimizes matched A/V/T/N bundles with $r_{\mathrm{act}}$/$r_{\mathrm{src}}$/$r_{\mathrm{ans}}$, transition consistency, weakest-condition protection, and risk-constrained GRPO. Multi-seed CMC-Bench holdout and Modality-Bias means lead on Acc(I$^\star$), Acc(N), ConfabR, CDR, $\Delta$Acc, and near-zero $B$ while keeping Acc(T$^\star$) strong; ablations show that removing $r_{\mathrm{trans}}$ or freezing duals reintroduces one-sided commits or over-deferral. Gains concentrate on visual-wins conflict, calibrated abstention, and modality balance, supporting counterfactual consistency as a training target for reliable multimodal arbitration.

\bibliography{references}
\end{document}